%% file: trustomni_mmasia_camera_ready.tex
\documentclass[sigconf]{acmart}

\usepackage{booktabs}
\usepackage{microtype}
\hypersetup{hidelinks}
\usepackage{array}
\usepackage{bm}

\AtBeginDocument{%
  }

\newif\ifpublicationpreview
\input{publication_metadata.tex}

\AtBeginEnvironment{itemize}{\emergencystretch=2em}

\newcommand{\method}{SafeRank-ReID}
\newcommand{\ours}{SafeRank-ReID}
\newcommand{\bestval}[1]{{\bfseries\boldmath\ensuremath{\bm{#1}}}}
\newcommand{\secondval}[1]{\underline{\ensuremath{#1}}}

\begin{document}

\flushbottom
\setlength{\parskip}{0pt plus 1pt}

\title{SafeRank-ReID: Candidate-Conditioned Re-Ranking under Identity Conflicts for Multimodal Person Re-Identification}

\author{Yizhi Wang}
\affiliation{%
  \institution{Hangzhou Dianzi University}
  \city{Hangzhou}
  \country{China}}
\email{24050537@hdu.edu.cn}

\author{Zixuan Li}
\affiliation{%
  \institution{Hangzhou Dianzi University}
  \city{Hangzhou}
  \country{China}}
\email{24050517@hdu.edu.cn}

\author{Zhuwen Chen}
\affiliation{%
  \institution{Hangzhou Dianzi University}
  \city{Hangzhou}
  \country{China}}
\email{24050924@hdu.edu.cn}

\author{Feiwei Qin}
\authornote{Corresponding author.}
\affiliation{%
  \institution{Hangzhou Dianzi University}
  \city{Hangzhou}
  \country{China}}
\email{qinfeiwei@hdu.edu.cn}


\begin{abstract}
Multimodal Person Re-Identification (ReID) exploits complementary identity cues from different sensing modalities to mitigate the performance degradation caused by challenging imaging conditions. Existing methods typically estimate Modality Reliability at the query level, making it difficult to identify conflicting inputs that retain normal visual quality but correspond to an incorrect identity. Such inputs may receive high confidence, promote irrelevant gallery candidates, and lead to negative fusion. To address this issue, we propose SafeRank-ReID, a candidate-conditioned safe re-ranking framework for Multimodal Person Re-Identification (ReID). To capture the varying support of the same modality for different candidates, we introduce a candidate-conditioned evidence arbitration mechanism that integrates unimodal retrieval evidence, input quality, and cross-modal consistency to estimate trust and conflict logits for each modality--candidate pair. Together, these logits contribute to the normalized modality weights, allowing support strength to vary across modalities and candidates while reducing the contribution of potentially unreliable evidence. A candidate-set refinement module subsequently applies self-attention to jointly model candidate representations and predicts residual scores from their competitive relations, thereby refining the initial fused ranking. For fusion results that remain uncertain after refinement, a query-level fallback gate estimates an interpolation coefficient from the overall conflict level and cross-modal uncertainty, adaptively combining the fused scores with those of the most reliable unimodal branch. Experiments on RGBNT201 and MSVR310 show that SafeRank-ReID maintains competitive clean-input retrieval while improving robustness to Identity Conflicts across person and vehicle benchmarks. 
\end{abstract}



\maketitle

\section{Introduction}
\label{sec:introduction}

Person ReID matches people across non-overlapping cameras and supports long-term visual search~\cite{zheng2015market,zheng2016mars,ye2022survey}. Although convolutional and transformer models have substantially improved single-image matching~\cite{sun2018pcb,wang2018mgn,zhou2019osnet,he2021transreid}, RGB observations remain unreliable under poor illumination, occlusion, and large viewpoint changes. Multimodal systems address this limitation by combining visible, near-infrared (NIR), and thermal-infrared (TIR) observations. The additional spectra often preserve identity cues that are weak or absent in RGB images~\cite{wu2017rgb,li2020xiv,chen2021nfs,park2021align}.

\begin{figure}
  \centering
  \IfFileExists{figures/motivation.pdf}{%
    \includegraphics[width=\columnwidth]{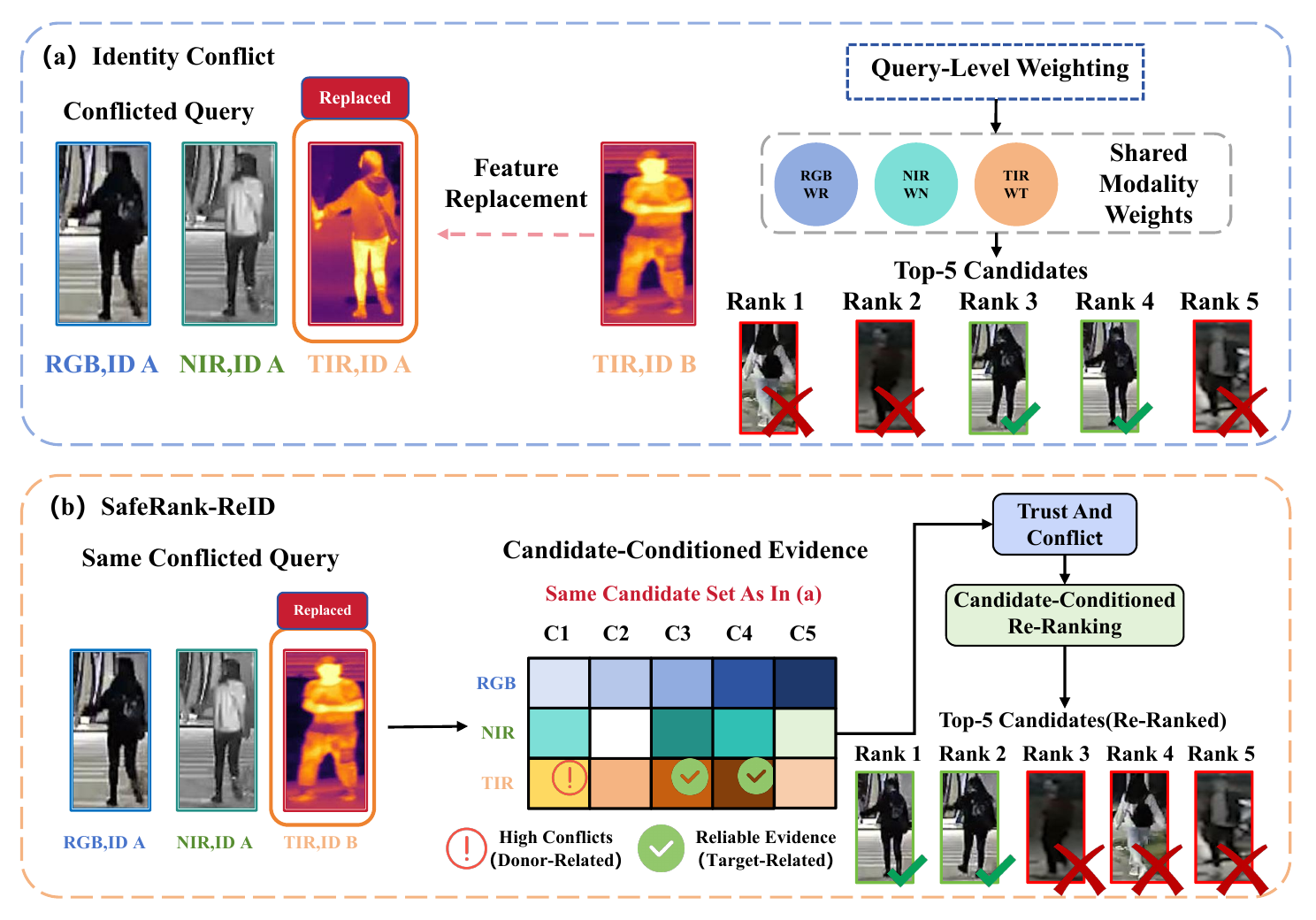}%
  }{%
    \fbox{\parbox[c][3.45cm][c]{0.94\columnwidth}{\centering
    Conflict query: RGB \textbar{} NIR \textbar{} TIR from donor identity\\[0.55em]
    Query-level weighting: Top-5 gallery ranking\\[0.45em]
    SafeRank-ReID: Top-5 gallery ranking\\[0.65em]
    \footnotesize Insert real query images, ranked gallery samples, and conflict scores.}}
  }
  \Description{A two-panel motivation figure. Query-level weighting assigns one shared weight to each modality and ranks a donor candidate highly after TIR replacement. SafeRank-ReID instead evaluates modality--candidate evidence and reranks the same candidates.}
  \caption{Motivation for \method{}. (a) Under an identity conflict, query-level weighting applies one shared weight per modality and may promote the donor identity. (b) \method{} estimates trust and conflict for each modality--candidate pair, then reranks the same candidates to reduce conflicting evidence. Green and red borders mark correct and incorrect matches.}
  \label{fig:teaser}
\end{figure}

Multimodal ReID has moved from separate modality streams to learned alignment and interaction. Early multispectral models either align heterogeneous features or use a separate branch for each modality \cite{zheng2021robust,wang2024htt}. Recent methods build on vision transformers and vision-language pretraining \cite{dosovitskiy2021vit,radford2021clip}, then use token selection, prompt tuning, mixture-of-experts, modality decoupling, and test-time adaptation to obtain stronger joint representations \cite{wang2024top,zhang2024magic,lin2025dmpt,wang2025idea,feng2025mfrnet,feng2025mdreid,liu2026signal,zheng2026proxyttt}. These advances mainly improve representation quality under correctly associated inputs. Their fusion decisions are usually shared across all gallery candidates or are made before retrieval, which implicitly assumes that a modality has one reliability level for the entire query.

This assumption is fragile when the input tuple contains an identity conflict. A tracking handoff, imperfect temporal alignment, or data-association error can attach a visually valid observation of another identity to an otherwise correct query. The corrupted modality may retain high image quality and a sharp retrieval distribution, so quality-based weighting can assign it high confidence. More importantly, its effect is candidate dependent: it may strongly support the incorrect donor identity while providing little evidence for other candidates. A single query-level weight cannot express this relation. Fig.~\ref{fig:teaser} contrasts this failure mode with candidate-conditioned arbitration over the same candidate set. Missing-modality training and robust fusion help when evidence is absent or degraded~\cite{neverova2016moddrop,ma2022missing,ma2023calibrating}, but they do not directly address plausible evidence from the wrong identity.
This failure mode suggests a different place for reliability estimation. Rather than deciding modality weights before observing retrieval candidates, the ranker can compare the evidence that each modality provides for each candidate. The union of independent Top-$K$ lists is a compact working set: it preserves candidates supported by any available branch while exposing rank agreement, score disagreement, and cross-modal support. Such evidence also permits a conservative decision when learned fusion becomes uncertain. A practical method should improve conflict robustness without sacrificing clean-input accuracy, operate with missing modalities, and still return a complete gallery ranking.

We introduce \method{}, a candidate-conditioned safe re-ranking framework. Each available modality first retrieves independently, and their Top-$K$ results form a candidate union. Candidate-Conditioned Evidence Arbitration predicts trust and conflict logits for every modality--candidate pair and uses them to reduce unsafe evidence before normalization. Candidate-Set Refinement then makes a bounded residual correction from competition within the shortlist. Finally, Reliable-Branch Fallback blends the refined scores with a learned unimodal branch when the fused evidence remains uncertain. The framework is trained with explicit identity replacement and a safety objective that compares the final ranking with observable fusion and unimodal references.

Our contributions are as follows:
\begin{itemize}
  \item We formulate identity-conflict robustness as a candidate-conditioned ranking problem and introduce Candidate-Conditioned Evidence Arbitration. Built on the union of independent Top-$K$ lists, it estimates trust and conflict for each modality--candidate pair and suppresses unsafe evidence before fusion.
  \item We introduce Candidate-Set Refinement for ambiguities that remain after local arbitration. A compact self-attention encoder compares candidates within the shortlist and applies bounded residual corrections without replacing the underlying retrieval geometry.
  \item We develop Reliable-Branch Fallback for uncertain fusion. It interpolates the refined scores with a learned deployable unimodal branch, while conflict, branch-selection, fallback, and safety objectives connect identity-conflict detection to harmful-fusion avoidance.
  \item Experiments on RGBNT201 and MSVR310 show that the full model largely preserves clean-input retrieval accuracy. Under identity conflicts, it improves mAP and Rank-1 and reduces NFR against controlled fusion rules and external methods evaluated under the same feature-level protocol. Ablations support the three mechanisms across person and vehicle retrieval.
\end{itemize}

\section{Related Work}
\label{sec:related}

\subsection{Multimodal Object Re-Identification}

Multimodal ReID initially addressed spectral discrepancy with shared embeddings and auxiliary cues~\cite{wu2017rgb,li2020xiv}, followed by explicit cross-modal alignment~\cite{chen2021nfs,park2021align}. With aligned RGB--NIR--TIR inputs, attention shifted toward cross-modal interaction. TOP-ReID rearranges modality tokens and Magic Tokens selects informative ones~\cite{zheng2021robust,wang2024top,zhang2024magic}, whereas prompt-, decoupling-, and expert-based methods separate shared identity evidence from modality-specific cues~\cite{lin2025dmpt,wang2025idea,li2025icpl,wang2025demo,feng2025mfrnet,feng2025mdreid}. UniCat offers a different view, showing that independently trained branches with late fusion remain competitive~\cite{crawford2023unicat}.

Recent work has made multimodal ReID more flexible at deployment. AIO and FlexiReID accept different modality subsets~\cite{li2024aio,sun2025flexireid}, while HTT and ProxyTTT adapt features at test time~\cite{wang2024htt,zheng2026proxyttt}. These methods address missing inputs or distribution shift, but they still assume that the modalities grouped into a query depict the same identity. If this association is wrong, a visually plausible modality can confidently favor an incorrect gallery candidate. Its evidence may agree with one candidate and conflict with another, so one reliability weight for the whole query is insufficient. RGBNT201 and MSVR310 use the same RGB--NIR--TIR interface for person and vehicle retrieval, respectively~\cite{zheng2021robust,zheng2023msvr310}, making it possible to evaluate the same association error across object categories.

\subsection{Reliable Fusion and Re-Ranking}

Reliability-aware fusion mainly addresses missing, degraded, or adversarial sources~\cite{yang2021single}. ModDrop and missing-modality training expose models to unavailable inputs~\cite{neverova2016moddrop,ma2022missing}; calibration and trusted aggregation estimate branch- or query-level confidence~\cite{han2021trusted,ma2023calibrating}. MMSF preserves modality-specific information and rebalances visible--infrared branches under image corruption~\cite{josi2025mmsf}. These methods associate unreliability with absence, visual degradation, or global uncertainty. Identity conflict differs because an input may appear normal while supporting the wrong identity, making its risk candidate dependent.

Classical ReID re-ranking revises an initial list through neighborhood consistency, as in $k$-reciprocal encoding and ECN~\cite{zhong2017reranking,sarfraz2018ecn}. Such methods rely on one similarity source or compatible rankings and cannot determine which modality provides unsafe evidence for a candidate. \method{} targets this gap after independent retrieval by arbitrating modality--candidate Top-$K$ evidence, refining shortlist competition, and falling back to a learned unimodal branch when fusion remains uncertain.
\section{Method}
\label{sec:method}

\begin{figure*}[t]
  \centering
  \includegraphics[width=\textwidth]{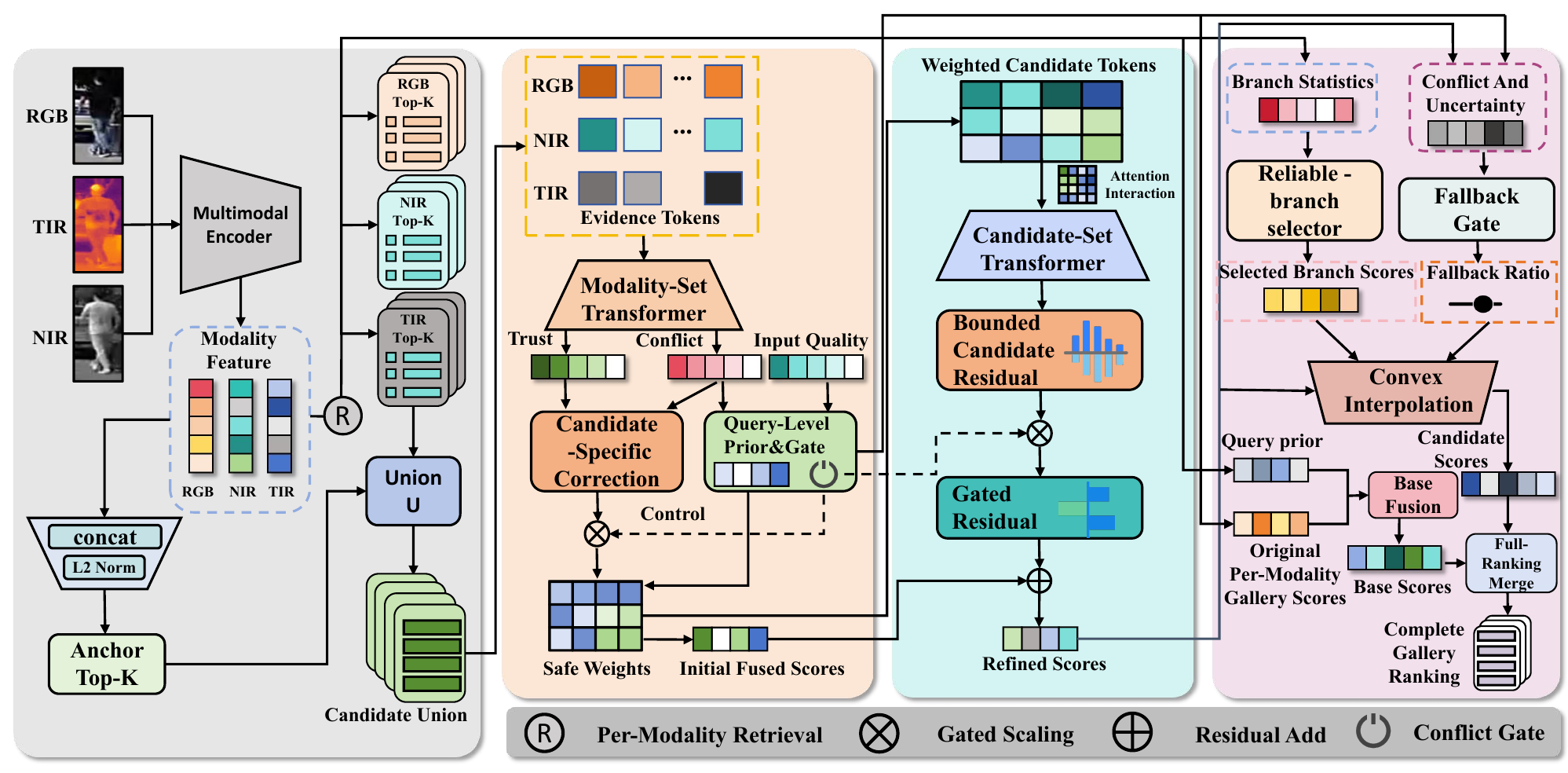}
  \Description{Pipeline from available RGB, NIR, and TIR query features through independent retrieval, candidate-conditioned arbitration, candidate-set refinement, reliable-branch fallback, and complete gallery ranking.}
  \caption{Overview of \method{}. Available modalities retrieve independently. Their Top-$K$ union is encoded as modality--candidate evidence, followed by conflict-aware arbitration, bounded candidate-set refinement, and Reliable-Branch Fallback. The full gallery order is recovered by merging reranked candidates with query-level base scores.}
  \label{fig:overview}
\end{figure*}

\subsection{Problem Setting and Conflict Model}

The complete ranking pipeline is summarized in Fig.~\ref{fig:overview}. Given an existing multimodal encoder, let $\mathcal{M}$ be the set of available modalities, with $M=|\mathcal{M}|$, and let $z_m^q\in\mathbb{R}^{d}$ and $z_{m,g}\in\mathbb{R}^{d}$ denote its normalized query and gallery features for modality $m$. The proposed ranker operates on these modality-specific representations and is independent of the encoder architecture. Each branch produces
\begin{equation}
s_{mg}=\langle z_m^q,z_{m,g}\rangle .
\end{equation}
An identity conflict occurs when one query feature is replaced by the same-modality feature of a different identity while the gallery remains unchanged. The replacement retains plausible visual content but breaks identity agreement within the query, separating this failure from ordinary image corruption.

\subsection{Candidate Union and Evidence Construction}

Each available modality retrieves independently. We also form an anchor descriptor $z_{\mathrm{anc}}$ by concatenating the available modality features and applying $\ell_2$ normalization, yielding $s_{\mathrm{anc},g}=\langle z_{\mathrm{anc}}^q,z_{\mathrm{anc},g}\rangle$. The working set is
\begin{equation}
\mathcal{C}_q=\bigcup_{m\in\mathcal{M}}\operatorname{TopK}(s_m)\ \cup\ \operatorname{TopK}(s_{\mathrm{anc}}).
\end{equation}
This union retains a candidate strongly supported by any branch. For candidate $j\in\mathcal{C}_q$ and modality $m$, we construct
\begin{equation}
x_{jm}=[s_{jm},\,\bar r_{jm},\,q_m,\,u_{jm},\,a_m,\,d_{jm},\,c_j],
\end{equation}
where $\bar r_{jm}$ is inverse normalized rank, $q_m$ is produced by a lightweight modality-specific quality head applied to $z_m^q$ and learned jointly with the ranker, $u_{jm}$ indicates Top-$K$ support, and $a_m$ indicates availability. The score deviation $d_{jm}$ measures disagreement with the available-modality mean, while $c_j$ is the fraction of available branches that retrieve $j$ in their Top-$K$ lists. Together, these terms describe retrieval strength, input quality, and cross-modal agreement without gallery labels.

\subsection{Candidate-Conditioned Evidence Arbitration}

A two-layer modality-set Transformer~\cite{vaswani2017attention} maps the evidence tokens of each candidate to a trust logit $\alpha_{jm}$ and a conflict logit $\ell_{jm}$. During training, the query--modality replacement label is broadcast to valid candidate tokens to supervise $\ell_{jm}$. We define the query-level posterior $p_m=\sigma(|\mathcal{C}_q|^{-1}\sum_j\ell_{jm})$. The residuals $\Delta\alpha_{jm}$ and $\Delta\ell_{jm}$ are obtained by subtracting the corresponding candidate mean and applying $\tanh$; they capture candidate-specific behavior and receive gradients from the ranking objectives. The safe weights are
\begin{equation}
w_{jm}=\operatorname{softmax}_{m}\!\left(b_m+g_q(\Delta\alpha_{jm}-\lambda\Delta\ell_{jm})\right),
\quad f_j=\sum_m w_{jm}s_{jm},
\end{equation}
where $b_m=q_m-\beta p_m$, $g_q=\max_{m\in\mathcal{M}}p_m$, and $\beta,\lambda\geq0$ control query- and candidate-level suppression. Centering preserves the query prior and bounding limits local corrections. Clean queries remain close to quality fusion, while a conflicting modality can be retained for one candidate and suppressed for another.

\subsection{Candidate-Set Refinement}

Arbitration evaluates the modalities of each candidate but does not compare candidates directly. We therefore pass the weighted candidate tokens through a two-layer candidate-set Transformer. A score head predicts a bounded residual and subtracts its mean over valid candidates, producing $\delta_j$. The refined score is
\begin{equation}
\tilde f_j=f_j+g_q\,\delta_j.
\end{equation}
The gate limits refinement when little correction is needed, while centering keeps the residual on the scale of the initial scores. The resulting update adjusts close candidates without replacing the initial retrieval geometry.

\subsection{Reliable-Branch Fallback and Full Ranking}

A reliability head first selects a deployable unimodal reference. For each available branch, it receives the quality estimate, leading score and margin, predicted non-conflict probability, score concentration, and top-candidate support. The supervision target is the branch with the lowest detached retrieval risk, and a straight-through hard selector produces its candidate scores $s_j^{\star}$.

The fallback gate summarizes the overall conflict level, candidate uncertainty, cross-modal score dispersion, and fusion-weight entropy to predict $\rho_q\in[0,1]$. It combines the refined and reference scores as
\begin{equation}
F_j=(1-\rho_q)\tilde f_j+\rho_q s_j^{\star}.
\end{equation}
Thus, uncertain fusion moves toward an observable unimodal reference rather than relying on a fixed fallback rule. To recover a complete ranking, let $\pi_m=\operatorname{softmax}_m(b_m)$ and define the query-level base score $B_g=\sum_m\pi_m s_{mg}$. The final gallery score is
\begin{equation}
S_g=\begin{cases}F_j,&g=j\in\mathcal{C}_q,\\B_g,&g\notin\mathcal{C}_q.\end{cases}
\end{equation}
Bounded residuals and convex fallback keep $F_j$ comparable with $B_g$; sorting $S$ therefore does not force every pooled candidate ahead of the gallery tail. The learned operations cover at most $(M+1)K$ candidates; missing modalities are masked throughout, and inference requires neither gallery identities nor conflict annotations.

\subsection{Training Objectives}

Training batches contain clean tuples, missing-modality masks, and identity replacements. We optimize
\begin{equation}
\mathcal{L}=\mathcal{L}_{\mathrm{ret}}+\lambda_c\mathcal{L}_{\mathrm{conf}}+
\lambda_r\mathcal{L}_{\mathrm{rel}}+\lambda_f\mathcal{L}_{\mathrm{fb}}+
\lambda_s\mathcal{L}_{\mathrm{safe}}.
\end{equation}
$\mathcal{L}_{\mathrm{ret}}$ is cross-entropy over the final candidate scores, and $\mathcal{L}_{\mathrm{conf}}$ is binary cross-entropy for the broadcast query--modality replacement labels. These labels supervise conflict detection, not candidate-level localization; candidate-specific suppression is learned through the retrieval and safety objectives.

$\mathcal{L}_{\mathrm{rel}}$ trains the reliable-branch selector. $\mathcal{L}_{\mathrm{fb}}$ chooses the lowest-risk target from interpolation coefficients $\{0,0.25,0.5,0.75,1\}$ between the refined and selected-branch rankings. Here, listwise risk is the negative log-probability of the positive under softmax-normalized candidate scores. 

Finally, $\mathcal{L}_{\mathrm{safe}}$ penalizes final listwise risk above the better observable reference, either query-level base fusion or the selected unimodal branch, and adds clean-query consistency with mean fusion. Candidate-level ranking losses are evaluated only when the positive occurs in $\mathcal{C}_q$; candidate coverage is reported in Section~\ref{sec:experiments}. These objectives connect conflict detection to the ranking decision used at inference.
\begin{table*}[t]
\caption{Main results at conflict rate $p=0.5$. Bold and underlined values denote the best and second-best result in each column, respectively; ties share the same style. All values are percentages, and lower NFR is better.}
\label{tab:main}
\centering
\scriptsize
\setlength{\tabcolsep}{1.8pt}
\renewcommand{\arraystretch}{1.03}
\resizebox{\textwidth}{!}{%
\begin{tabular}{@{}llccccc@{\hspace{4pt}}ccccc@{}}
\toprule
& & \multicolumn{5}{c}{RGBNT201} & \multicolumn{5}{c}{MSVR310}\\
\cmidrule(lr){3-7}\cmidrule(l){8-12}
& & \multicolumn{2}{c}{Clean} & \multicolumn{3}{c}{Conflict} & \multicolumn{2}{c}{Clean} & \multicolumn{3}{c}{Conflict} \\
\cmidrule(lr){3-4}\cmidrule(lr){5-7}\cmidrule(lr){8-9}\cmidrule(lr){10-12}
Method & Source & mAP & R1 & mAP & R1 & NFR $\downarrow$ & mAP & R1 & mAP & R1 & NFR $\downarrow$\\
\midrule

Mean fusion & --
& \secondval{82.50{\pm}0.00} & $85.29{\pm}0.00$ & $62.48{\pm}0.53$ & $61.56{\pm}0.37$ & $39.00{\pm}0.32$
& \bestval{58.41{\pm}0.00} & \secondval{73.27{\pm}0.00} & $44.90{\pm}0.38$ & $53.92{\pm}0.85$ & $64.92{\pm}0.10$\\
Quality-weighted & --
& \bestval{82.75{\pm}0.78} & \bestval{85.89{\pm}1.14} & $62.76{\pm}0.50$ & \secondval{62.20{\pm}1.06} & $38.68{\pm}1.09$
& $56.85{\pm}1.33$ & $73.21{\pm}0.68$ & $44.38{\pm}1.81$ & $55.72{\pm}1.13$ & $65.48{\pm}3.21$\\
Query-level & --
& $82.45{\pm}0.03$ & $85.29{\pm}0.00$ & \secondval{66.36{\pm}0.56} & $61.56{\pm}0.37$ & \secondval{33.98{\pm}0.21}
& $56.89{\pm}2.49$ & $72.02{\pm}2.15$ & \secondval{45.41{\pm}1.57} & $55.78{\pm}4.02$ & $64.24{\pm}1.60$\\
\midrule

MFRNet & ICML'25~\cite{feng2025mfrnet}
& 74.62 & 76.32 & $54.64{\pm}0.30$ & $53.67{\pm}0.38$ & $46.81{\pm}1.74$
& 48.06 & 63.79 & $37.20{\pm}0.26$ & $45.57{\pm}0.68$ & $66.55{\pm}2.00$\\
MDReID & NeurIPS'25~\cite{feng2025mdreid}
& 75.92 & 79.78 & $54.46{\pm}0.82$ & $53.99{\pm}1.00$ & $50.40{\pm}1.34$
& 49.59 & 66.67 & $38.46{\pm}0.49$ & $48.05{\pm}1.99$ & $65.82{\pm}0.45$\\
ICPL-ReID & TMM'26~\cite{li2025icpl}
& 70.72 & 73.56 & $51.15{\pm}0.92$ & $50.20{\pm}0.81$ & $52.87{\pm}1.04$
& 55.55 & \bestval{76.31} & $44.24{\pm}0.41$ & \secondval{57.36{\pm}0.78} & $68.98{\pm}0.76$\\
Signal & AAAI'26~\cite{liu2026signal}
& 75.99 & 78.47 & $55.59{\pm}0.51$ & $54.39{\pm}0.80$ & $45.10{\pm}0.67$
& 52.76 & 70.73 & $40.71{\pm}0.26$ & $51.61{\pm}1.22$ & \secondval{63.45{\pm}1.67}\\
ProxyTTT & AAAI'26~\cite{zheng2026proxyttt}
& 82.44 & 84.93 & $60.48{\pm}0.69$ & $59.53{\pm}0.81$ & $42.22{\pm}1.18$
& \bestval{58.41} & \secondval{73.27} & $44.91{\pm}0.42$ & $53.98{\pm}0.94$ & $64.86{\pm}0.20$\\
\midrule
\textbf{\ours{}} & \textbf{Ours}
& $82.13{\pm}0.32$ & \secondval{85.49{\pm}0.28} & \bestval{69.35{\pm}1.19} & \bestval{71.39{\pm}1.56} & \bestval{29.17{\pm}2.17}
& \secondval{57.79{\pm}0.41} & $72.90{\pm}0.61$ & \bestval{49.42{\pm}0.72} & \bestval{64.62{\pm}0.45} & \bestval{57.36{\pm}2.48}\\
\bottomrule
\end{tabular}
}
\end{table*}
\section{Experiments}
\label{sec:experiments}

\subsection{Datasets, Protocols, and Metrics}

\noindent We evaluate whether candidate-conditioned re-ranking improves robustness to identity conflicts without materially weakening clean-input retrieval, whether this behavior persists across conflict rates and available-modality patterns, and which components account for the improvement. The study includes controlled fusion rules with shared branch features and external methods evaluated under a common feature-level protocol.

\noindent\textbf{RGBNT201.} RGBNT201 is a multispectral person ReID benchmark containing aligned RGB, NIR, and TIR observations of 201 identities~\cite{zheng2021robust}. We use the official train/test split and camera-aware retrieval protocol.

\noindent\textbf{MSVR310.} MSVR310 contains aligned RGB, NIR, and TIR observations of 310 vehicles captured under varied viewpoints and environmental conditions~\cite{zheng2023msvr310}. We use the official identity split. This benchmark tests whether the ranking mechanism transfers to another object category; it is not treated as additional evidence for person-specific appearance modeling.

\noindent\textbf{Identity Conflict Protocol.} We use three pre-generated manifests. At conflict rate $p$, each selects a proportion $p$ of queries, samples one eligible modality uniformly, and replaces its representation with the same modality from a different identity. The gallery remains clean, and the donor is drawn from the same camera when labels permit. The manifests are shared by all methods within each run; $p=0.5$ unless stated otherwise. Quality-weighted fusion, Query-level weighting, and \ours{} are trained with three random seeds and paired with the corresponding manifests, so their mean $\pm$ sample standard deviation includes training and conflict-sampling variation. Mean fusion is deterministic on clean inputs, and its conflict deviation comes only from the manifests. Each external method uses one fixed checkpoint per dataset, so its deviation also reflects conflict sampling only.

\noindent\textbf{Metrics.} We report mean average precision (mAP), Rank-1 accuracy, and negative fusion rate (NFR). For query $q$, let $AP_q^{\mathrm{fuse}}$ denote the fused average precision and $AP_{qm}$ the average precision of available modality $m$. NFR is the proportion of queries satisfying $AP_q^{\mathrm{fuse}}<\max_m AP_{qm}$, so a lower value means that fusion harms fewer queries. Conflict F1 uses a fixed posterior threshold of 0.5, and AUROC measures threshold-free separation; both are computed over query--modality labels.

\subsection{Implementation and Comparison Protocol}

For the controlled comparison, we extract 768-dimensional branch features with a trained ProxyTTT ViT-B/16 encoder and optimize only the ranking modules. Images are resized to $256\times128$ on RGBNT201 and $128\times256$ on MSVR310, consistent with the aspect ratios used in their public protocols. We set the candidate size to $K=50$. Both set encoders have a hidden width of 256, four attention heads, two layers, and a dropout rate of 0.1. We optimize the ranker with AdamW for 60 epochs on RGBNT201 and 20 epochs on MSVR310, using a batch size of 32, an initial learning rate of $3\times10^{-4}$, and cosine decay. The conflict, reliability, fallback, and safety loss weights are 0.5, 0.2, 0.2, and 0.5, respectively, with a safety slack of 0.02. The identity-replacement and missing-modality probabilities during training are 0.50 and 0.15, respectively.

The controlled baselines use the same cached branch features as \ours{}. Mean fusion averages the available scores, Quality-weighted fusion assigns one quality-derived weight to each branch, and Query-level weighting predicts one branch weight for the complete query. This comparison isolates the ranking rule from the feature extractor. We train MFRNet~\cite{feng2025mfrnet}, MDReID~\cite{feng2025mdreid}, ICPL-ReID~\cite{li2025icpl}, Signal~\cite{liu2026signal}, and ProxyTTT~\cite{zheng2026proxyttt} from their public implementations using the recommended configurations. For a common feature-level comparison, we replace one pre-fusion query-modality representation according to the same conflict manifests and evaluate clean and conflicted retrieval with a shared evaluator. The external block therefore reports performance under this common protocol rather than each method's native inference pipeline.

\begin{figure}[!h]
  \centering
  \includegraphics[width=\columnwidth]{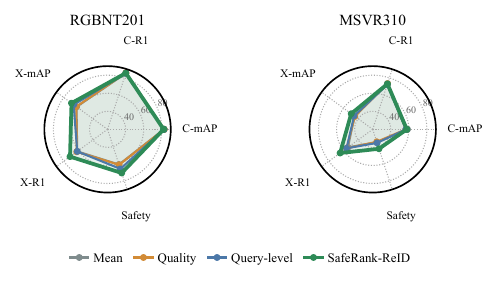}
  \Description{Two radar charts compare controlled fusion methods on clean mAP, clean Rank-1, conflict mAP, conflict Rank-1, and fusion safety for RGBNT201 and MSVR310.}
  \caption{Controlled performance profiles. C and X denote clean and conflict conditions; Safety is $100-\mathrm{NFR}$. Table~\ref{tab:main} gives exact values.}
  \label{fig:radar}
\end{figure}

\begin{table*}[!t]
\caption{Core ablation at conflict rate $p=0.5$, using seed 42 and the same fixed conflict manifest. Bold and underlined values denote the best and second-best result in each dataset column, respectively; ties share the same style. Higher is better except for NFR. Conflict Mechanism denotes conflict supervision and pre-normalization suppression.}
\label{tab:ablation}
\centering
\scriptsize
\setlength{\tabcolsep}{2.4pt}
\renewcommand{\arraystretch}{1.05}
\resizebox{\textwidth}{!}{%
\begin{tabular}{@{}>{\centering\arraybackslash}p{3.5cm}ccccc@{\hspace{7pt}}ccccc@{}}
\toprule
& \multicolumn{5}{c}{RGBNT201} & \multicolumn{5}{c}{MSVR310}\\
\cmidrule(lr){2-6}\cmidrule(l){7-11}
Variant & mAP & R1 & NFR$\downarrow$ & F1 & AUROC & mAP & R1 & NFR$\downarrow$ & F1 & AUROC\\
\midrule
\textbf{Full \ours{}} & \bestval{69.77} & \bestval{71.41} & \bestval{29.78} & \bestval{73.32} & \bestval{93.57} & \bestval{49.50} & \secondval{64.47} & \bestval{56.68} & \secondval{63.58} & \bestval{92.78}\\
w/o Evidence Arbitration & 63.17 & 66.39 & 38.28 & 52.48 & 86.88 & 47.78 & 63.28 & \secondval{57.19} & 60.14 & \secondval{91.72}\\
w/o Conflict Mechanism & 58.54 & 63.04 & 45.45 & 26.69 & 49.60 & 41.64 & 53.98 & 68.87 & 28.69 & 44.93\\
w/o Set Refinement & 67.54 & 64.72 & 34.61 & \bestval{73.32} & \bestval{93.57} & 48.16 & 63.65 & 57.67 & \secondval{63.58} & \bestval{92.78}\\
w/o Branch Fallback & \secondval{68.51} & \secondval{69.74} & \secondval{30.14} & \bestval{73.32} & \bestval{93.57} & \secondval{48.72} & \bestval{64.67} & 57.48 & \secondval{63.58} & \bestval{92.78}\\
w/o Safety Loss & 65.49 & 65.31 & 36.96 & \secondval{61.29} & \secondval{92.49} & 42.68 & 55.50 & 67.34 & \bestval{67.29} & 91.64\\
\bottomrule
\end{tabular}%
}
\end{table*}

\subsection{Comparative Experiments}

We compare clean retrieval and identity-conflict robustness under two complementary settings in Table~\ref{tab:main}. The shared-feature block directly tests the ranking design because only the fusion or re-ranking rule changes. The external-method block extends the comparison to complete systems under the same feature-level corruption and retrieval protocol. Separating the two settings distinguishes improvements from the ranker from those due to different feature extractors.

On RGBNT201, \ours{} remains within 0.62 mAP of the best shared-feature result on clean inputs. Under conflicts, it exceeds Query-level weighting by 2.99 mAP and 9.83 Rank-1 points and reduces NFR by 4.81 points. Compared with ProxyTTT, the strongest external result under the common protocol, the gains reach 8.87 mAP and 11.86 Rank-1 points, together with a 13.05-point NFR reduction. The contrast between the small clean difference and the larger conflict gains is consistent with the intended role of \ours{}: it changes how contradictory evidence is handled rather than replacing the underlying representation.

MSVR310 shows a similar trade-off. Mean fusion is 0.62 mAP and 0.37 Rank-1 points higher on clean inputs, while \ours{} improves conflict mAP by 4.01 points and conflict Rank-1 by 8.84 points over Query-level weighting. It also surpasses the best external conflict result by 4.51 mAP and 7.26 Rank-1 points and lowers NFR by at least 6.09 points. The gain on a second object category suggests that candidate-conditioned ranking is not tied to person retrieval; claims about person-specific appearance remain confined to RGBNT201.

The controlled profiles in Fig.~\ref{fig:radar} show the clean--robustness trade-off. \ours{} stays close to the controlled baselines on clean inputs while improving the conflict-oriented axes, most clearly conflict Rank-1 and fusion safety.

\subsection{Conflict Strength and Missing Modalities}

\begin{figure}[t]
  \centering
  \includegraphics[width=\columnwidth]{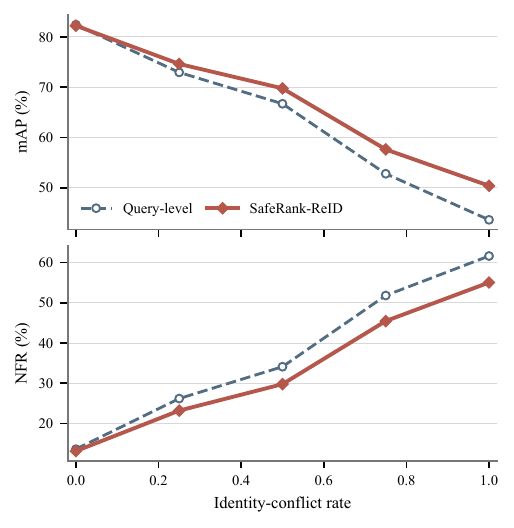}
  \Description{Line plots of mAP and negative fusion rate for SafeRank-ReID and query-level weighting from zero to full identity-conflict rate.}
  \caption{RGBNT201 robustness as the identity-conflict rate increases. Each method uses its fixed seed-42 ranker checkpoint, and both use the same encoder features and conflict manifests.}
  \label{fig:conflict-rate}
\end{figure}

The conflict-rate sweep in Fig.~\ref{fig:conflict-rate} shows how ranking behavior changes as conflicts become more frequent. At $p=0.25$, \ours{} exceeds Query-level weighting by 1.68 mAP and lowers NFR by 2.99 points. At $p=1.0$, the gaps widen to 6.78 mAP and 6.58 NFR points. This trend is consistent with candidate-conditioned arbitration: its advantage becomes clearer when a query-level reliability value is more likely to conceal candidate-specific contradictions. With no replacement, the mAP values remain close, at 82.24 for \ours{} and 82.43 for Query-level weighting.

\begin{figure}[t]
  \centering
  \includegraphics[width=\columnwidth]{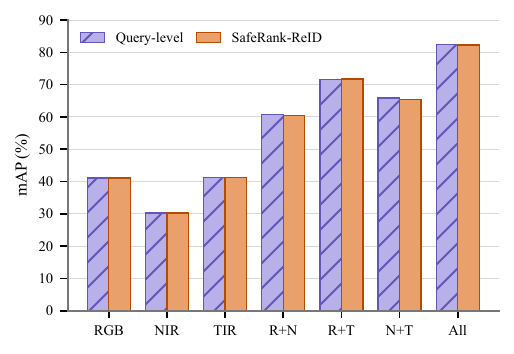}
  \Description{Grouped bars compare SafeRank-ReID and query-level weighting for each available RGB, NIR, and TIR subset on RGBNT201.}
  \caption{RGBNT201 mAP under available-modality subsets. Single-modality rows coincide because no fusion is performed.}
  \label{fig:missing}
\end{figure}

We next test whether candidate-conditioned conflict handling incurs an additional cost when modalities are unavailable (Fig.~\ref{fig:missing}). With a single available modality, \ours{} and Query-level weighting coincide because fusion is inactive. Relative to Query-level weighting, the mAP changes for RGB+NIR, RGB+TIR, and NIR+TIR are $-0.29$, $+0.16$, and $-0.43$ points, respectively; the change with all modalities is $-0.19$. The absolute gap therefore remains within 0.43 points across all multimodal subsets. These small differences show that the proposed mechanism preserves the behavior of the underlying fusion baseline when evidence is missing; its primary role is to handle contradictory rather than absent evidence. At $K=50$, positive-candidate coverage is 100.00\%/98.92\% on clean/conflict RGBNT201 and 99.15\%/98.70\% on MSVR310. Candidate omission is therefore not the main source of the remaining retrieval errors.

\subsection{Ablation Study}

We assess each component through ablations on RGBNT201 and MSVR310. The results in Table~\ref{tab:ablation} show that the full model provides the strongest overall balance between retrieval accuracy and fusion safety. Removing the Conflict Mechanism causes the largest and most consistent drop, including near-random AUROC, while removing Evidence Arbitration also degrades retrieval, NFR, and conflict detection on both datasets. The large drops from these two removals show that candidate-conditioned conflict handling accounts for most of the robustness gain. Set Refinement and Branch Fallback act after conflict prediction, so their contribution appears in the retrieval metrics rather than F1 or AUROC. The full model improves mAP and NFR over the variant without fallback on both datasets, and the Safety Loss consistently improves mAP, Rank-1, and NFR. The downstream components therefore convert conflict-aware evidence into safer final rankings.

\section{Conclusion}

This paper presents \method{} for multimodal person re-identi\-fication, which assesses the reliability of modality evidence at the candidate level to mitigate negative fusion caused by identity association errors. By combining independent retrieval results, input quality, and cross-modal consistency, the method adjusts modality weights for different candidates to reduce the influence of conflicting evidence. Building on this, the model refines the ranking through relationships among candidates and falls back toward a selected reliable unimodal branch when fusion remains uncertain. Experiments on RGBNT201 show that the method improves person retrieval accuracy under identity conflicts and reduces the negative fusion rate while maintaining competitive clean-input retrieval performance; separate training and evaluation on MSVR310 also support its applicability to vehicle re-identi\-fication. Although the method still depends on candidate-set coverage and the current feature-replacement protocol does not capture the temporal structure of real association errors, future work could improve candidate selection and evaluate identity association errors in continuous observations to further strengthen its robustness in real-world settings.
\bibliographystyle{ACM-Reference-Format}
\bibliography{references}

\end{document}

%% file: publication_metadata.tex
\publicationpreviewfalse
\copyrightyear{2026}
\acmYear{2026}
\setcopyright{cc}
\setcctype{by}
\acmConference[MMAsia '26]{ACM Multimedia Asia}{December 15--18, 2026}{Hanoi, AA, Vietnam}
\acmBooktitle{ACM Multimedia Asia (MMAsia '26), December 15--18, 2026, Hanoi, AA, Vietnam}
\acmDOI{10.1145/3862739.3863281}
\acmISBN{979-8-4007-3133-4/2026/12}